\documentclass[sigconf]{acmart}  % TODO use this for generating the camera ready version (all submissions)

\usepackage{color}
\usepackage{balance}
\usepackage{algorithm} 
\usepackage{algpseudocode} 

\def\BibTeX{{\rm B\kern-.05em{\sc i\kern-.025em b}\kern-.08emT\kern-.1667em\lower.7ex\hbox{E}\kern-.125emX}}

\renewcommand\footnotetextcopyrightpermission[1]{}
\copyrightyear{2023}
\acmYear{2023}
\setcopyright{acmlicensed}
\acmConference[xxxx 2023]{xxxx xxxx xxxx}{xxxx xxxx, 2023}{xxxx, xxxx}
\acmBooktitle{xxxx xxxx xxxx xxxx ,xxxx, 2023, xxxx, xxxx}
\begin{document}

%
% The "title" command has an optional parameter, allowing the author to define a "short title" to be used in page headers.
\title{Covert Communication Based on the Poisoning Attack in Federated Learning}
% \title[Short title (shown in page headers)]{Full title}

%
% The "author" command and its associated commands are used to define the authors and their affiliations.
% Of note is the shared affiliation of the first two authors, and the "authornote" and "authornotemark" commands
% used to denote shared contribution to the research.
\author{Junchuan Liang}
\email{liangjunchuan@stu.sicnu.edu.cn}
\author{Rong Wang}
\authornotemark[1]
\email{rwang@sicnu.edu.cn}
\affiliation{%
  \institution{School of Computer Science, Sichuan Normal University}
  \streetaddress{Chenlong Street 1819}
  \city{Chengdu}
  \state{China}
  \postcode{610101}
}

% \author{Rong Wang}
% \affiliation{%

% \affiliation{%

% \affiliation{%

% \affiliation{%

% \affiliation{%



%
% Use this command to specificy the authors in the page headers.
% Assuming the the name of the first authors is Ben Trovato, then: 
% - if there is only one author, use B. Trovato
% - if there are more authors, use B. Trovato et al. 
% \renewcommand{\shortauthors}{B. Trovato et al.}

%
% The abstract is a short summary of the work to be presented in the article.
\begin{abstract}
Covert communication has become an important area of research in computer security. It involves hiding specific information on a carrier for message transmission and is often used to transmit private data, military secrets, and even malware. In deep learning, many methods have been developed for hiding information in models to achieve covert communication. However, these methods are not applicable to federated learning, where model aggregation invalidates the exact information embedded in the model by the client. To address this problem, we propose a novel method for covert communication in federated learning based on the poisoning attack. Our approach achieves 100\% accuracy in covert message transmission between two clients and is shown to be both stealthy and robust through extensive experiments. However, existing defense methods are limited in their effectiveness against our attack scheme, highlighting the urgent need for new protection methods to be developed. Our study emphasizes the necessity of research in covert communication and serves as a foundation for future research in federated learning attacks and defenses. 
\end{abstract}

%    \todo{use this command to highlight remaining work.}

% \ccsdesc[100]{Model~Network reliability}

%
% Keywords. The author(s) should pick words that accurately describe the work being
% presented. Separate the keywords with commas.

%
% A "teaser" image appears between the author and affiliation information and the body
% of the document, and typically spans the page.
%   \includegraphics[width=\textwidth]{figures/sample-teaser}
%   \Description{Enjoying the baseball game from the third-base seats. Ichiro Suzuki preparing to bat.}

%
% This command processes the author and affiliation and title information and builds
% the first part of the formatted document.
\maketitle

%
% This command processes the content.tex file
\section{Introduction}
Covert communication \cite{106,107,108} is a method of transmitting secret information has been receiving a lot of attention. Pony \cite{114} and Glupteba \cite{115} have been observed to use bitcoin transactions for transmitting messages. And StegoNet \cite{116} successfully embeds malware into the artificial intelligence (AI) model. Numerous studies \cite{113} have been made about using models as carriers in deep learning. However, there are few studies on covert communication in federated learning.

Federated learning \cite{1} is a decentralized machine learning that allows models to be trained data across multiple clients without the need to share the raw data. This technique has been applied in various fields (such as finance \cite{4}, healthcare \cite{2,3}, and the Internet of Things (IoT) \cite{6}) and has the potential to shape the future of artificial intelligence in a privacy-obsessed world \cite{8}. The process of federated learning involves initializing a global model, local training of the model on each client's data, aggregation of the updated model parameters, and evaluation of the model's performance. This process is repeated for multiple iterations to improve the model's performance. %However, as the client-side training process and data are invisible, poisoning attacks \cite{9,35,36} pose an unavoidable risk.

The focus of this paper is on achieving covert communication between two clients in federated learning using only model aggregation. Federated learning is a preferred choice in institutions such as hospitals and banks that prioritize privacy protection to obtain better models. In such settings, models are only transmitted to a fully trusted server, which carefully reviews them before performing model aggregation and distributing the aggregated models to clients. In this hypothetical scenario, we assume that an attacker is a compromised software whose goal is to leak important data with 100\% accuracy from a client to a receiver posing as a benign participant. However, this process will face three main problems.

\textbf{Issue 1:} The first issue an attacker needs to consider is that the client's important data is physically isolated, except for the model that will be sent to a fully trusted server. Therefore, the attacker needs to find a way to hide the information into the model.

\textbf{Issue 2:} The second issue is that the server performs a rigorous review of the models submitted by the client, including model similarity comparison, accuracy detection, differential privacy, etc. The attacker needs to hide the information while ensuring that the model appears normal to pass the review process.

\textbf{Issue 3:} The third issue that the attacker needs to consider is that the server performs federated aggregation after receiving the client's model, such as FedAvg. This means that the attacker needs to ensure the accuracy of the hidden information in the model after recalculation, to prevent distortion or loss of the covert communication.

Previous methods proposed in \cite{113} and \cite{116} have demonstrated the ability to hide information within a model while maintaining its functionality. However, these methods are not suitable for federated learning scenarios due to the server's model aggregation process, which invalidates the embedded information. The backdoor-based approach proposed by \cite{43} enables information transfer between federated learning clients but is limited to transferring only 1 bit of data, making it unsuitable for our scenario.

\begin{figure}[h]
  \centering
  \includegraphics[width=\linewidth]{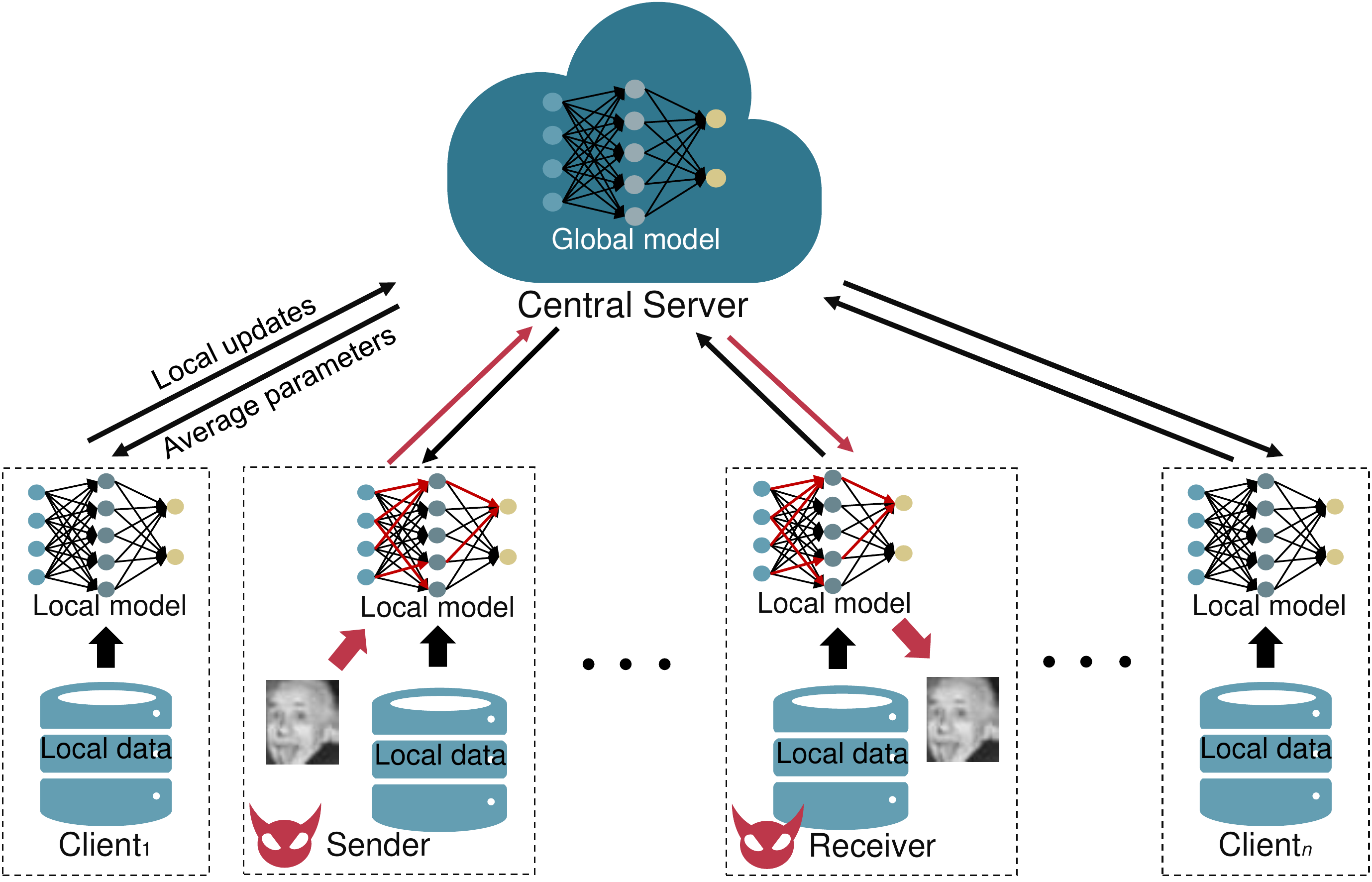}
  \caption{Federated learning covert communication (The red arrows represent the message that the sender wants to transmit.)}
  \Description{Federated learning covert communication. }
  \label{fig1}
\end{figure}

To address the limitations of existing methods, we propose a novel approach to federated learning covert communication (Figure \ref{fig1}) based on poisoning attacks  \cite{9,35,36}. This approach is motivated by two factors. Firstly, the invisibility of client training data and processes in federated learning makes it vulnerable to poisoning attacks  \cite{9}, making our proposed technique highly relevant. Second, our repeated poisoning technique significantly increases communication capacity, achieving 10,000-bit data transfer compared to the 1-bit capacity of backdoor-based approaches  \cite{43}.

% This paper proposes using poisoning techniques to investigate the possibility of implementing covert communication \cite{111,112,106,10} between two clients during federated learning (Fig.\ref{fig1}). Previous studies have mostly focused on directly embedding information in a carrier \cite{116}, while in federated learning, the client's model is the carrier of information. However, since the client model is recomputed at the server, directly embedding information can fail. Through experiments, we find that by modifying the values of the client-side model weights, the corresponding weights of the global model can be shifted in the direction of the modification. By utilizing this offset, covert information communication between clients can be achieved.

Our approach not only enables the accurate transmission of large amounts of information but also circumvents the careful scrutiny of the server. We evaluated the latest approaches to the federated learning of poisoning attack defense in recent years, such as differential privacy \cite{17,18}, FLAME \cite{41}, and GAA \cite{39}. Even if the central server uses a validation dataset to detect the clients' updates \cite{103}. We found that these methods do not prevent covert communication. This suggests that research into relevant defense options is imminent.

The paper contributes to a better understanding of the security implications of covert communication in federated learning and paves the way for further research in this area. The contributions of this paper are three-fold.

\begin{itemize}
\item we propose a novel approach to achieve federated learning covert communication using the poisoning attack. Our method can successfully achieve $100\%$ accurate transmission of information between two federated learning clients only through the federated learning aggregation process.
\item we propose an optimization method for our covert communication scheme, which enables more covert and larger capacity information transmission. The communication capacity of our scheme is also evaluated theoretically.
\item Experiments demonstrate the accuracy, stealthiness, and robustness of our approach. We analyze the effectiveness of current defense methods against our attack and point out new possible defense schemes.
\end{itemize}

\section{Background}
\subsection{Federated Learning and Poisoning attack}
Federated learning is a decentralized machine learning technique that allows multiple clients to train a global model without the need for data sharing. The key idea is to update the global model by aggregating the local updates from each client, which preserves the privacy of client data. The implementation process of federated learning typically involves the following steps.

Step 1. A central server distributes the initial global model to a set of clients.

Step 2. Each client device trains the model on its local data and sends the updated model back to the central server.

Step 3. The central server aggregates the model updates from all clients to create a new global model.

Step 4. Steps 2 and 3 are repeated for a certain number of rounds until a predetermined stopping criterion is reached.

However, federated learning can be vulnerable to various security threats, such as the poisoning attack. Poisoning attacks \cite{9,35,36} in federated learning involve a malicious client intentionally sending malicious data to the server. Such malicious data can lead to inaccurate or biased results. 

In recent years, a variety of poisoning attacks in federated learning have been proposed, including data poisoning \cite{36}, and model poisoning \cite{103}. And it is pointed out that poisoning attacks are inevitable due to the invisibility of the client training process and data in \cite{9}.

\subsection{Covert Communication Attack}
Covert communication attacks in \cite{106,107,108} involve using a communication channel to transmit information. Covert communication attacks are also known as covert channel attacks in \cite{108,109,110,111,112}. It is a major concern in the security of various systems, including cloud computing and pervasive computing systems. There are two main classes of covert channels, i.e., timing channels in \cite{112} and storage channels in \cite{108,109}. Timing channels rely on modulating resources to convey covert messages, while storage channels transfer information by storing values of inaccessible objects. Both types of attacks aim to make covert messages indistinguishable from legitimate traffic to avoid detection. Covert channels have been proposed in various communication protocols, including IP and TCP in \cite{110}. In wireless networks, timing channels are particularly challenging to implement due to the regularity of sampling and predictable processing. Overall, covert channel attacks can be powerful tools for industrial espionage and sabotage in \cite{107}. 

In this paper, we focus on the use of the poisoning attack to achieve covert communication between clients. This involves embedding covert messages within the model that can be decoded by the special client.

\section{Proposed Attack Method}
\subsection{Challenges}

To achieve covert communication between two clients in federated learning, we propose a poisoning attack-based strategy that embeds messages into the local model updates. Specifically, we first select a malicious client who participates in the training process, and we carefully craft their model updates to encode the covert messages. Then, we let these malicious clients send the poisoned model updates to the server, which will in turn distribute them to all other clients. In this process, we will face the following challenges.

\textbf{Challenge 1: How to send a message to the receiver?} One of the main challenges in covert communication is how to send a message from the sender to the receiver. In federated learning, all information embedded in the model by the message sender is recalculated by the server, which can result in the sender not being able to send the exact value to the receiver. Moreover, it is also challenging for the receiver to accurately distinguish the bits of information sent by the sender from the model updates. 

\textbf{Challenge 2: How to ensure communication covertly?} Another main challenge in covert communication is how to ensure that the hidden message is not detectable or distinguishable from normal model updates. To send information to the receiver, in the case of joint learning, the information can only be carried through the model. Therefore, the challenge is how to ensure that the model that hides the information looks the same as the normal model. 

\textbf{Challenge 3: How to ensure the accuracy of the global model and local model?} The third critical challenge in covert communication is to prevent any disruption or compromise to the federated learning process. Since our hidden communication relies on the federated learning process, it is essential to ensure that the process is executed correctly as a prerequisite for the existence of covert communication. At the same time, ensuring the accuracy of the local model can reduce the possibility of the sender being detected by the server.

\subsection{Attacker Model}
We define the following attacker's scenario and the attacker's knowledge, as well as the target of the attack.

\textbf{Attack scenario.} Our attack scenario is designed for a horizontal federated learning setting. In this setting, the central server collects models from clients and computing updates. Meanwhile, the clients train their respective models. In this scenario, the attacker participates in the federated learning process by disguising themselves as benign client. 

\textbf{Attackers.} To make the covert communication process more intuitive, we divide the attacker into two roles, i.e., the message sender and the message receiver. Both of them join the federated learning system disguised as benign clients. The sender is responsible for encoding a message into the model weights at a pre-agreed position, while the receiver is tasked with decoding the information from the global model based on the recorded and averaged weights over several rounds. The success of the task is determined by the accurate matching of the decoded information to the sent information.

\textbf{Attackers’ knowledge.} The sender and receiver can agree in advance on the position of the weights used to send messages in the model, based on their knowledge of the model architecture. The sender and the receiver can agree in advance on the way of encoding the content of the communication, based on their shared knowledge of the encoding scheme and the decoding algorithm. 

\subsection{Attack Method}
In federated learning, clients send their model updates to the server for aggregation and obtain the updated global model. In a poisoning attack, the attacker aims to inject malicious model updates that will manipulate the global model in a specific way. However, due to server model aggregation, it is difficult for the sender to send the exact values to the receiver.

To overcome this challenge, the sender can control the weight of the model update and make it closer to the desired direction. For example, if the sender wants to send a message that corresponds to the binary value of 0, the sender can adjust the weights of its model updates to be closer to the negative side of the scale; if the sender wants to send a message that corresponds to the binary value of 1, the sender can adjust the weights to be closer to the positive side of the scale. The proposed attack method is presented in Algorithm 1 and further explained as follows.

\renewcommand{\algorithmicrequire}{\textbf{Input:}}
\renewcommand{\algorithmicensure}{\textbf{Output:}}

\renewcommand{\thealgorithm}{1} %这里用来定义算法1，算法2等
    \begin{algorithm}
    \label{al1}
        \caption{Covert communication in federated learning} %标题
        \begin{algorithmic}[1] %每行显示行号，1表示每1行进行显示
            \Require The agreed position $w_p$ for encoding the message, total cycle $S_N$ and which one has $n$ rounds, and the Information encoded as binary $B = b_1, b_2, ..., b_{S_N}$ to be encoded.
            \Ensure The received message $R_B$ after decoding.
            \For{$S_0 \to S_N$} 
                % \State $C_1 \Leftarrow \emptyset, C_2 \Leftarrow \emptyset$ 
                \For{$S_i = 1,2,...,n$}     
                    % \State $w_{pl} = b_i$
                    \If {$sender$}
                    \State  $w_{pl} = b_i$; \Comment{Information embedded}
                    \State  The other weights are trained normally and the 
                    \State updated model is sent to the server;
                    \Else
                    \State Each client trains the model on their local data and 
                    \State sends the updated model to the server;
                \EndIf
                \State Server aggregates the model updates from all clients 
                \State and sends the new global model to each client;
                \State Receiver records the $w_{pg}$ and stores it; 
                \EndFor
                % \State $\tilde{u_1} \Leftarrow \frac{1}{\vert C_1 \vert}\sum_{x \in C_1} x$, $\tilde{u_2} \Leftarrow \frac{1}{\vert C_2 \vert}\sum_{x \in C_2} x$ 
            % \Endfor   
            \EndFor
            \State The receiver calculates the mean $w_{pg}$ every cycle; 
            \For{$S_0 \to S_N$}    \Comment{Information removal}
                \If{$mean (w_{pg}) > 0$} 
                    \State $R_B[s] = 1$;
                    \Else
                    \State $R_B[s] = 0$;
                \EndIf
            \EndFor
            \State \Return $R_B$
        \end{algorithmic}
    \end{algorithm}

As shown in Algorithm 1, $w_p$ is the weight that the sender and receiver agree to use to transmit the message. The $w_{pl}$ is the sender's local weight, which is the same location and size as $w_p$. And the $w_{pg}$ is the global weight, which is the same location and size as $w_p$.

Lines 1-2. The message is sent in cycles of $S_N$, and $n$ is the total round in $s_i$. 

Lines 3-4. The sender embeds the message into the model weights agreed on in advance. This step is repeated in one cycle.

Lines 5-6. The sender uses the remaining weights to train the model.

Lines 8-9. Benign client training model.

Lines 11-12. The server aggregates the model updates from all clients and sends the new global model to each client. 

Lines 13-14. The receiver records the weight value $w_p$ of the agreed position p from the global model and stores it.

Lines 16-24. The receiver calculates the mean $w_{pg}$ and decodes the message bits as follows. In one cycle if  mean $w_{pg} > 0$, the bit is set to 1, otherwise, it is set to 0.

The decoded message is then compared with the original bit sequence to obtain the received message $R_B$. 

\begin{figure}[h]
  \centering
  \includegraphics[width=\linewidth]{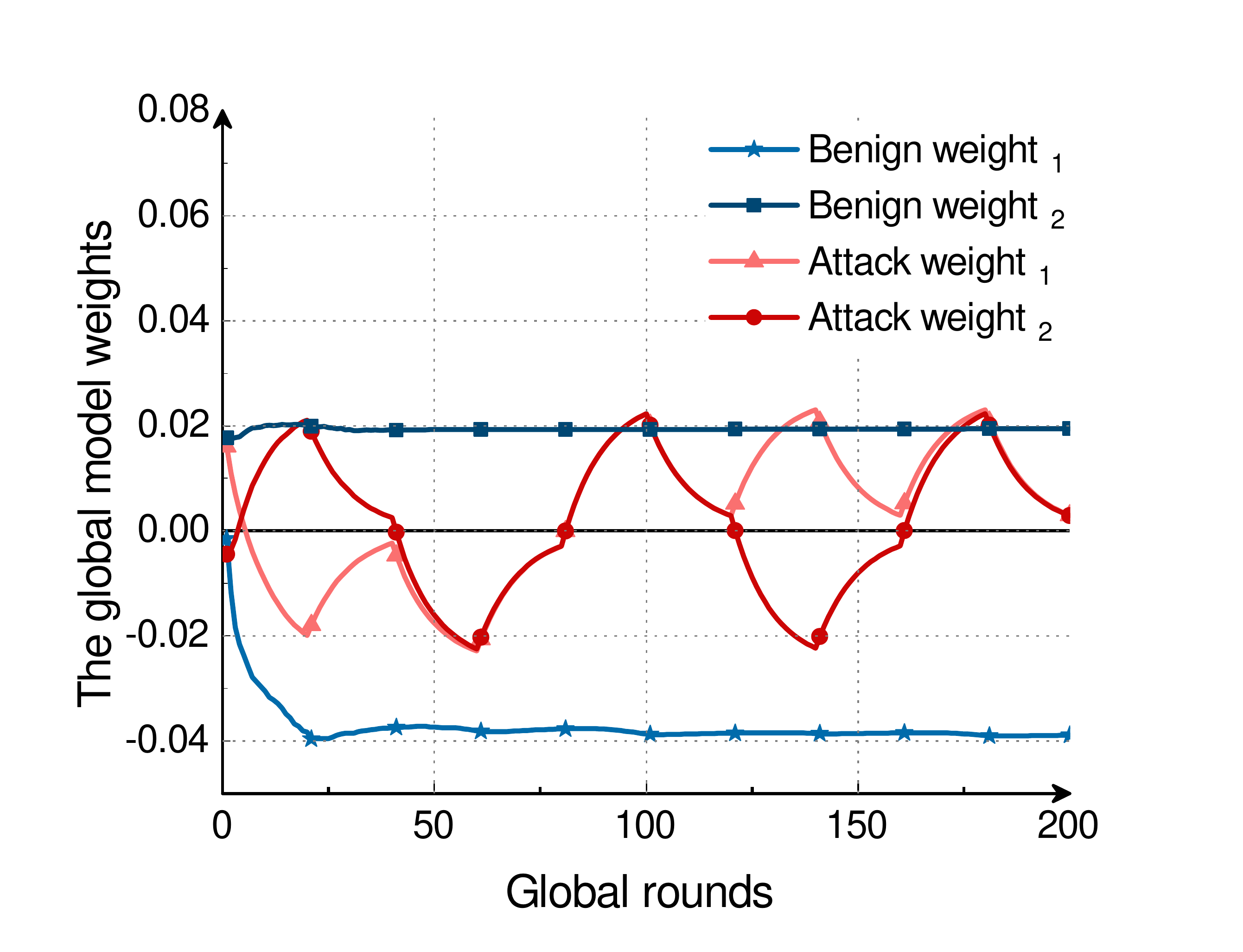}
  \caption{The variation of selected weights $w_{pg}$ in the global model.}
  \Description{The variation of selected weights in the global model. }
  \label{fig3}
\end{figure}

By controlling the direction of the weight updates in this way, the sender can embed their message in the model updates and have it transmitted through the server to the receiver shown as in Figure \ref{fig3}.

This process allows for covert communication between two clients in federated learning through the poisoning attack, where the message bits are encoded in specific model weight and transmitted through the server difficult to be detected by other clients or the server. The mean of the recorded position-specific weights is used to decode the message, and the message capacity can be controlled by the number of rounds and the number of positions used for encoding the bits. As shown in Figure \ref{fig3}, the blue line is the normal weight change curve, while the red line is the weight change curve used to transmit the information. The receiver gets the bit value by calculating the sum of the vertical coordinates of the curve in one cycle. If the value of the vertical coordinate in this cycle is greater than 0, then its bit value is 1, and otherwise, it is 0. It should be noted that the two red curves here are only 2 weights.

\subsection{More Covert Communication}
We propose the following techniques to increase the difficulty of detecting covert communication between two clients in federated learning.

\textbf{Random selection of weights.} Instead of using a fixed position in the model weights to encode the message bits, the sender and the receiver can randomly select the weights for the first round. Note, however, that the random selection here is chosen jointly by the message sender and the receiver in advance at random. That is, the position of the weights of these transmitted messages does not change during the training process. Attention is paid to the fact that the upper limit of the number of weights chosen should have a small impact on the model performance.

\textbf{Multiplication of factor.} To further mask the poisoned data and make it indistinguishable from benign data, the sender can multiply a factor when modifying the weights during poisoning. We propose using the root mean square $(RMS)$ of 500 randomly selected weight values as the factor. It captures the variation in the weight distribution and reduces the distinguishability between the poisoned and benign data.

\textbf{Zeroing back of weights.} Randomly initialized weights can introduce noise and interfere with the covert communication signal. To mitigate this, we propose zeroing back the weights to their initial values of 10 rounds (or another number of rounds) before sending the signal, so that the signal is less affected by the random initialization and the communication is more covert.
By using these techniques, covert communication can be further improved in terms of stealthiness and resilience to detection, while maintaining a sufficient communication capacity and minimizing the impact on the federated learning process.

\section{Analysis of Proposed Attack}
\subsection{Higher Communication Capacity}
To achieve greater communication capacity in covert communication between two clients in federated learning, we propose several improvement methods as follows. These methods can be used in combination or individually.

\textbf{Using more weights.} The current neural network parameters are often large and using only one weight to encode the message bits can limit the communication capacity. By selecting more weights for each round of poisoning, more bits can be transmitted over a longer time. For example, using 1000 weights and 20 rounds per cycle, we can send up to 10,000 bits of data in 200 rounds. This shows that adding more weights can increase communication capacity and improve covert communication performance. Note that the number of weights selected should not significantly affect the model performance.

\textbf{Faster sending frequency.} The transmission speed can be doubled by decreasing the number of rounds per cycle from 20 to 10. The number of rounds can be so small that only 1 round is needed. If you send 2,000 bits per cycle, you can send 200,000 bits in 100 cycles when the cycle is equal to 1 round.

\textbf{Better reception mechanism.} Instead of comparing the weight value at the agreed position with 0 to determine the bit of the accepted data, we can use the average value of the recorded position-specific weights as the comparison threshold. This has the advantage of being more covert, and more accurate. 

By using these techniques, the communication capacity of covert communication attacks in federated learning can be significantly increased, while maintaining a low detection rate and a high level of security. 

\subsection{Covert Channel Bandwidth}

To calculate the channel bandwidth for secret communication in joint learning, we need to consider several factors such as the number of weights used for transmission and the transmission cycle. Based on the assumptions provided, we can estimate the channel bandwidth as follows.

We use 1,000 weights to transmit the message, and each weight can successfully transmit 1 bit every 20 rounds. This means that in one cycle of 20 rounds, we can transmit 1,000 bits of information.

If we assume that the training process in federated learning lasts for $T$ rounds, then we can transmit $T/20$ cycles of information during the training process.

Therefore, the total amount of information that can be transmitted covertly during the training process is $B = (T/20) \times1000$ bits.
The channel bandwidth is the rate at which we can transmit this information over time and is given by the formula $R = B/T$, where $R$ is the channel bandwidth in bits per round.

Plugging in the above values, we get $R = (T/20)\times1000/T = 50$ bits per round.

This means that, on average, we can transmit 50 bits of covert information per round of training in federated learning, using the given assumptions. In the same 200 rounds, we can achieve 10,000-bit information transmissions.

%, which is much larger than the 1-bit information transmission in \cite{43}. Note that if fewer rounds are used, as well as more weights, the theoretical bandwidth is also greatly increased.

\section{Experiments}
\subsection{Evaluation Metrics}
In our evaluation of the proposed covert communication method, we use the following four metrics.

\textbf{Model accuracy.} The accuracy of the federated learning model is an important factor to consider when evaluating the effectiveness of a covert communication attack. We consider both the global model accuracy and local model accuracy in our evaluation. The global model accuracy is the accuracy of the model tested by the server after aggregating all client models, while the local model accuracy is the accuracy of each client model tested by the server. We assume that the server has a validation dataset in \cite{25,39,40} and could test each uploaded local model accuracy and global model accuracy at every round. By comparing the accuracy of the model before and after the covert communication process, we can evaluate the impact of covert communication on model accuracy.

\textbf{Stealthiness.} The goal of covert communication is to make the communication undetectable. Therefore, measuring the degree to which the attacker's communication is covert is an important evaluation metric. We use the L2-norm-based in \cite{38} approach and the cosine similarity in \cite{25,41} to evaluate the weights similarity between the attacker and the benign clients. The L2-norm-based approach calculates the Euclidean distance between two sets of weights, while the cosine similarity measures the angle between the two weight vectors. By comparing the L2-norm or cosine similarity between the attacker and benign clients, we can evaluate the level of stealthiness. 

\textbf{Robustness.} The ability to resist noise-based and perturbation-based defenses [15, 16] is an important factor to consider when evaluating the effectiveness of convert communication attacks. If a covert communication method is easily disrupted by noise or defenses, it may not be a viable technique for real-world applications. We assume that the server adds noise to the model uploaded by the attacker in each round and evaluates the level of noise using $N_l$. If $N_l$ is equal to 0, it means that there is no additional random Gaussian noise in the model parameters. If $N_l$ is equal to 1, it means that the model parameters are all random Gaussian noise. By measuring the robustness of our covert communication method, we can assess its ability to resist noise and perturbation-based defenses.

\textbf{Successful receive time.} The time the receiver successfully receives all the covert messages sent by the attacker is an important evaluation metric. If the covert communication process takes too long or is too unreliable, then it may not be a viable technique for real-world applications. By measuring the successful receive time, we can assess the efficiency and reliability of our covert communication method.

Overall, these evaluation metrics help us assess the effectiveness, efficiency, and robustness of our proposed covert communication method and compare it with other existing methods in the literature.

\subsection{Experimental Setup}

\textbf{Experimental data set and network model.} We use the handwriting recognition data set MNIST \cite{23} and the image recognition data set CIFAR10 \cite{22} to complete our experiments. MNIST is a handwriting recognition data set that includes 60,000 training samples and 10,000 test samples. Each sample is a $28\times28$ pixel handwritten number. And it has 10 class labels from the number “0” to the number “9”. CIFAR10 has 50,000 training samples and 10,000 test samples. Each sample is a $32\times32\times3$ image, such as “airplane”, “car”, etc. It also has 10 classes. The network model we used is a 2-layer convolution neural network and an 18-layer ResNet18 \cite{7}.

%We use MNIST as the training data for the 2-layer convolution neural network and CIFAR10 as the training data for ResNet18.

 \textbf{Sending data.} The sending data is the data that the attacker tries to send to the receiver. It includes a short text message and an image message in our experiments. The text message is: \textit{Do not answer! Do not answer!} The image message is a gray image of Einstein's head with a size of 11,040 bits. We send the text message in the 2-layer convolution neural network FL framework with 60 weights every round. And we send the image message using the ResNet18 with 2,250 weights every round.

\subsection{Experimental Results}

\begin{figure}[h]
  \centering
  \includegraphics[width=\linewidth]{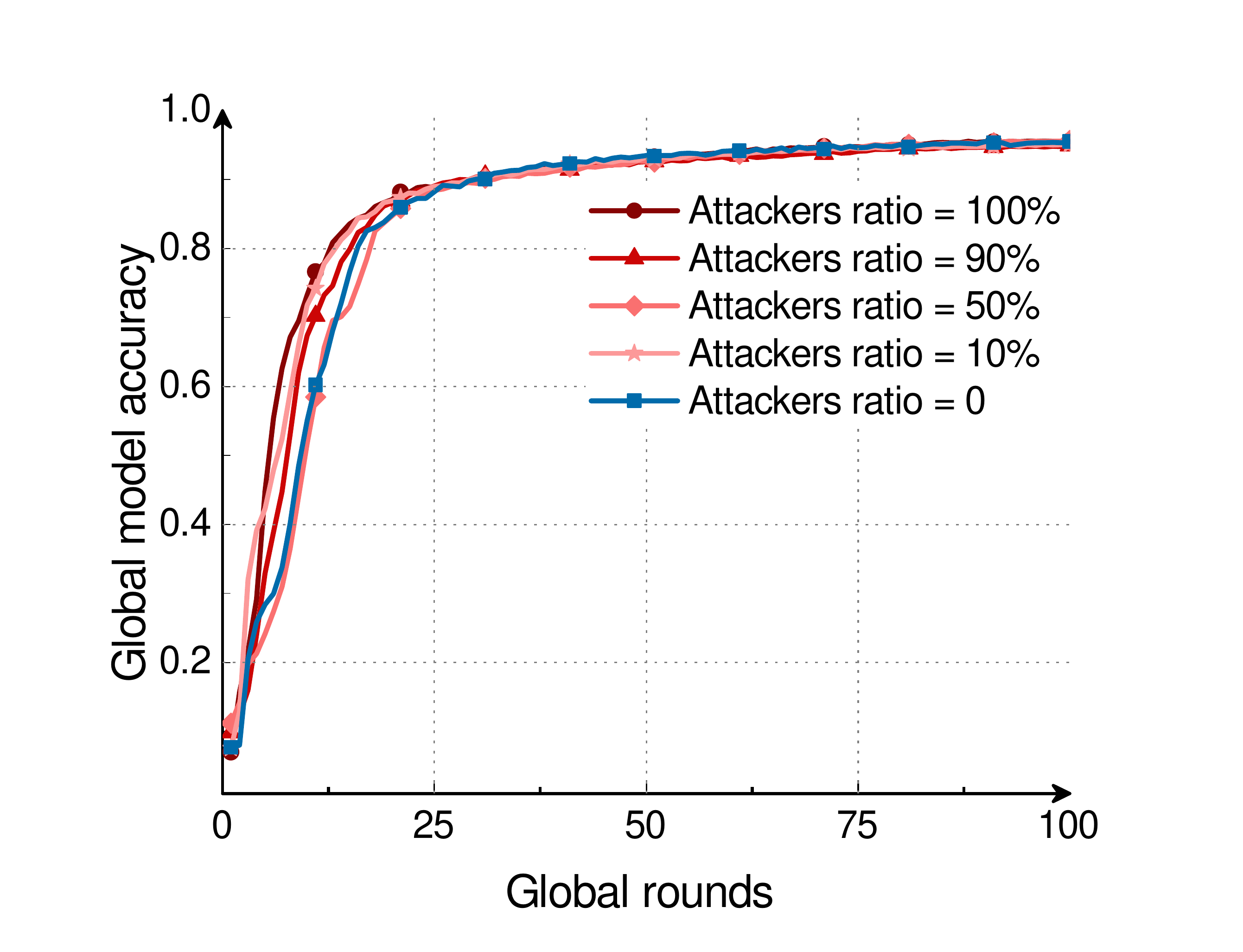}
  \caption{Global model accuracy at different attacker ratios.}
  \Description{Global model accuracy at different attacker ratios.}
  \label{fig4}
\end{figure}

\textbf{Our attack does not destroy the aggregation of the model.} As shown in Figure \ref{fig4}, we experimented with the variation of the global model accuracy for the different attacker ratios: 0\%, 10\%, 50\%, 90\%, and 100\%.  It can be seen from Figure \ref{fig4} that the different attacker ratios do not damage the convergence, even if all clients are attackers (100\%). This is where our attack method differs from destructive attacks (such as untargeted attacks \cite{35,36,37,38}). The reason why our approach does not damage the accuracy of the global model is that the attacker modifies only a small number of weights, and only a small number of weights changed has little effect on the accuracy of the model. This is because deep neural networks are highly redundant and have many parameters that are not critical to their performance. Additionally, many deep learning models are robust to small perturbations, making them less susceptible to attacks that modify only a small number of weights. Conversely, the attacker can modify the single weight very little, and these small modifications have little impact on deep neural networks.

\begin{figure}[h]
  \centering
  \includegraphics[width=\linewidth]{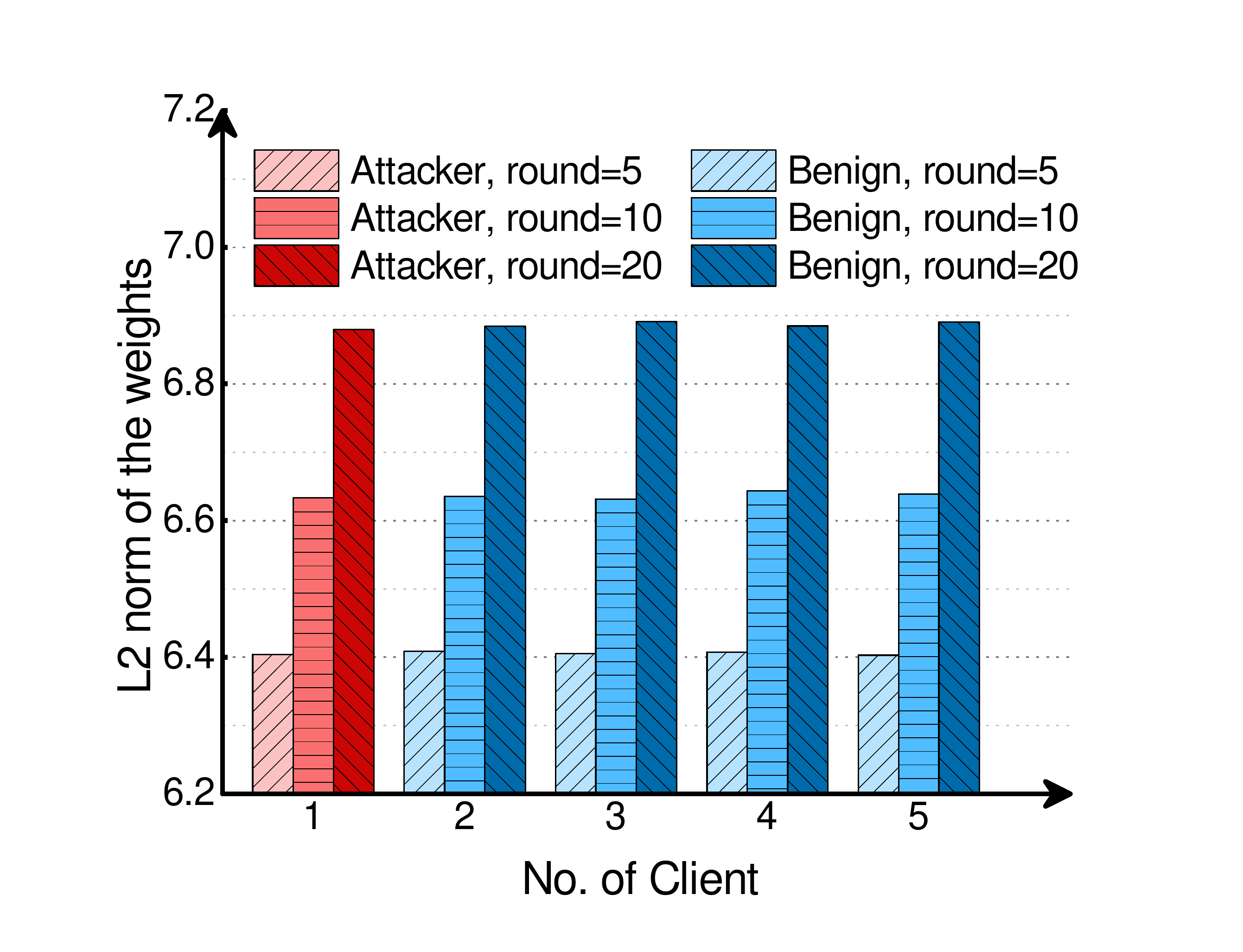}
  \caption{Comparison of L2 norm for attacker model and benign client model.}
  \Description{Comparison of L2 norm for attacker model and benign client model.}
  \label{fig5}
\end{figure}

\begin{figure}[h]
  \centering
  \includegraphics[width=\linewidth]{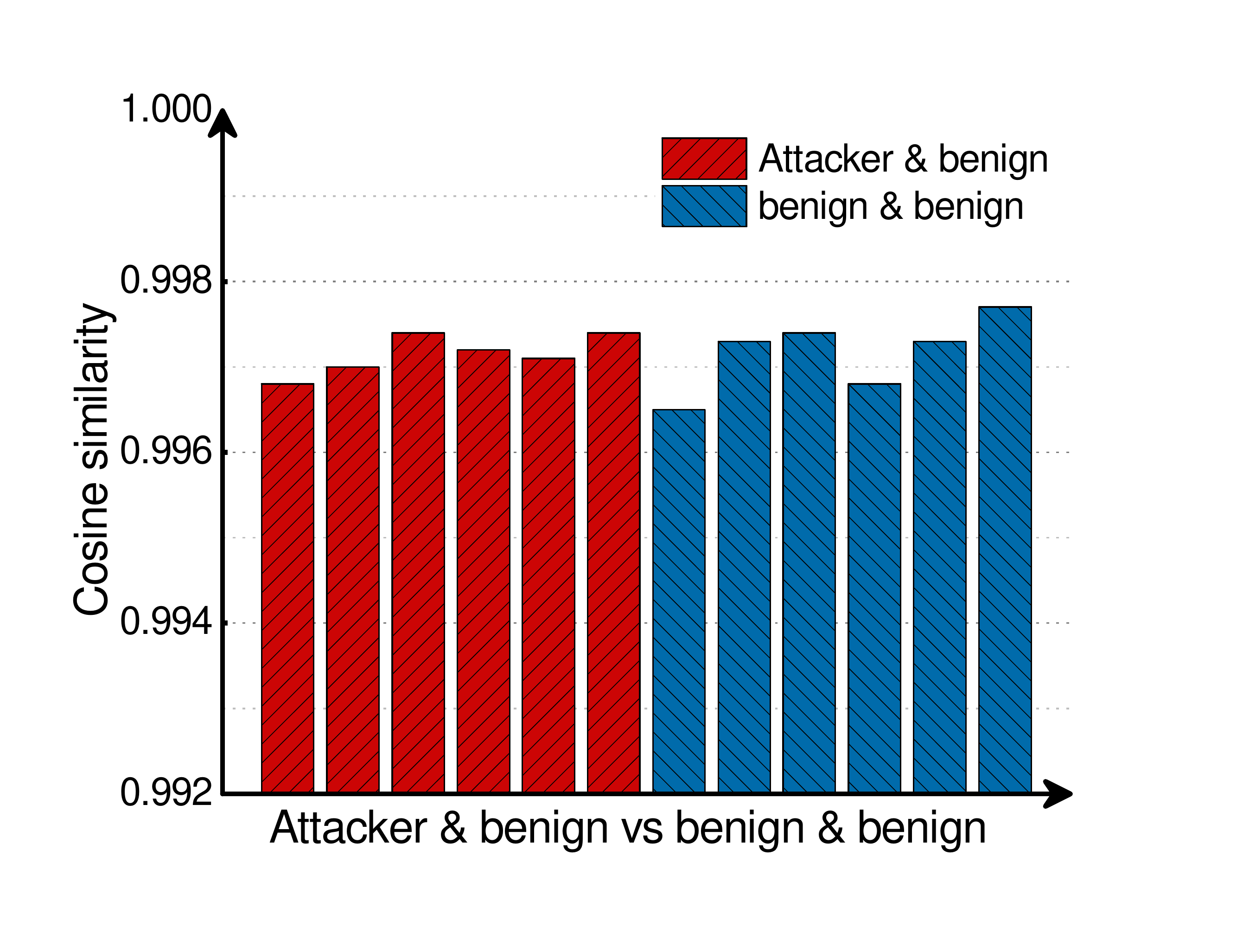}
  \caption{Cosine similarity between attacker and benign client, and cosine similarity between benign client and benign client.}
  \Description{Cosine similarity between attacker and benign client, and cosine similarity between benign client and benign client.}
  \label{fig6}
\end{figure}

\begin{figure}[h]
  \centering
  \includegraphics[width=\linewidth]{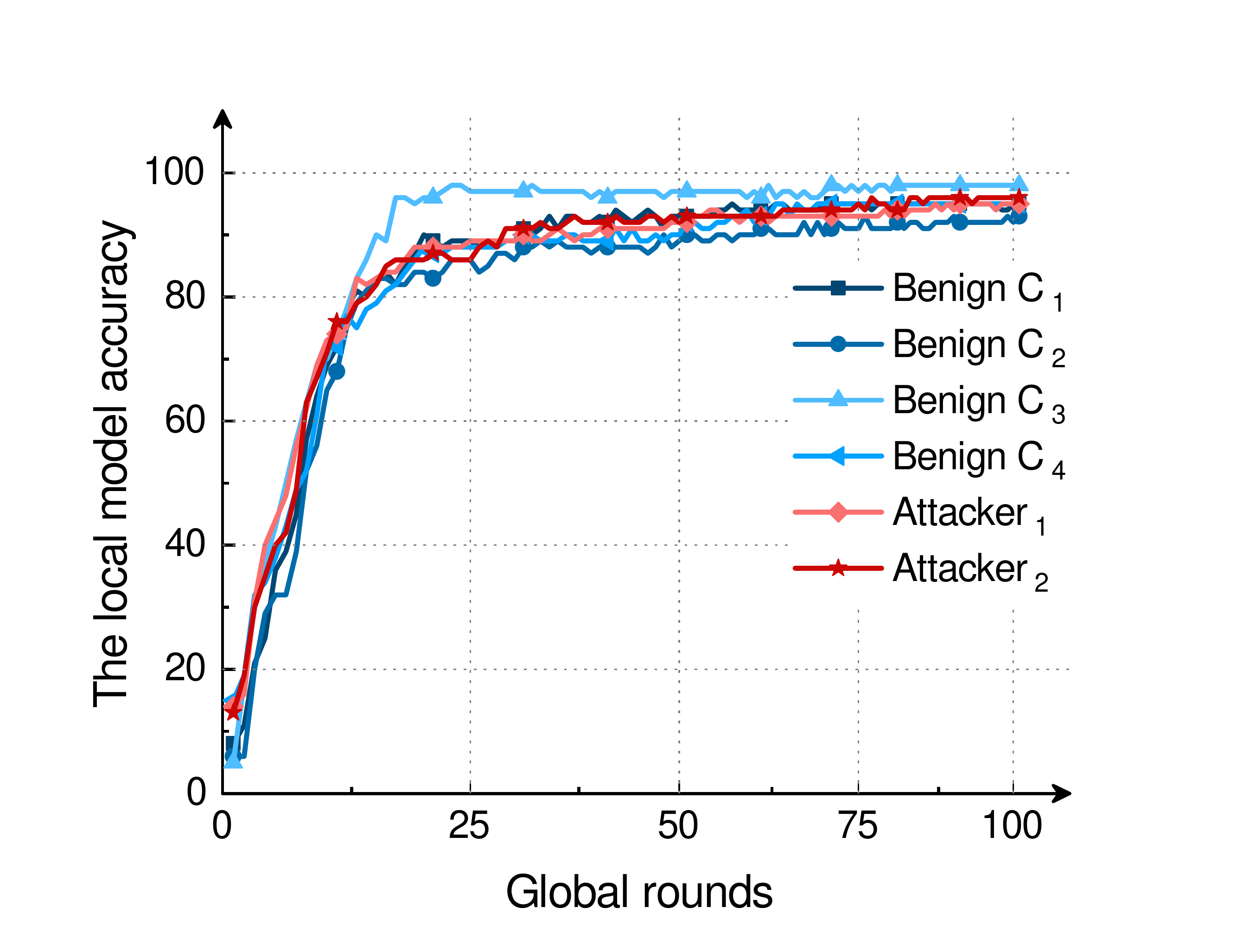}
  \caption{Comparing local model accuracy of attackers with benign clients.}
  \Description{Comparing local model accuracy of attackers with benign clients.}
  \label{fig7}
\end{figure}

\textbf{Our attack is stealthy.} We assume that the server could track each model uploaded by clients in each round and carefully examine each uploaded model. The norm-based approach \cite{38} has been shown very powerful in recent defenses against federated learning. The underlying idea of this approach is based on the assumption that the attacker and benign client models are inherently different. Figure \ref{fig5} provides a comparison of the $L2$ norm of the attacker and benign client models at different rounds. Figure \ref{fig5} reveals that the first client represents the attacker, while the other clients are benign. The vertical coordinates represent the weight $L2$ parameters. It can be observed that there are no notable differences between the weight $L2$ parameters of the attacker's weights and the other benign clients' weights at different stages of the federated learning process (i.e., round = 5, round = 10, and round = 20).

Recent work \cite{25,41} using the cosine similarity to detect attackers in federated learning exhibits strong defensive capabilities. So, we compare the cosine similarity between the attackers and the benign clients under our approach and the cosine similarity between the benign clients themselves. As shown in Figure \ref{fig6}, the cosine similarity between the attacker and the benign clients is close to 1, which means that the similarity between the attackers and the benign clients is high. Meanwhile, the cosine similarity between the attackers and the benign clients is very close to those of the benign clients. This means that the server cannot use cosine similarity to identify attackers from benign clients.

There are also federated learning defenses \cite{39,40} that assume that the server has validation data sets to test the accuracy of the local model uploaded by each client. We record the accuracy of the attacker and other benign clients in each round as shown in Figure \ref{fig7}. As in the figure, the attacker’s local mode accuracy is similar to those of the benign clients. This means it is hard for the server with a validation data set to detect the attacker. Moreover, the attacker is trained locally as the benign client, so it has the same positive effect on federated learning. This situation of the attacker continues throughout the federated learning process. Behavior-based detection is ineffective against our attacks because our attack does not destroy or control the model.

\begin{figure}[h]
  \centering
  \includegraphics[width=\linewidth]{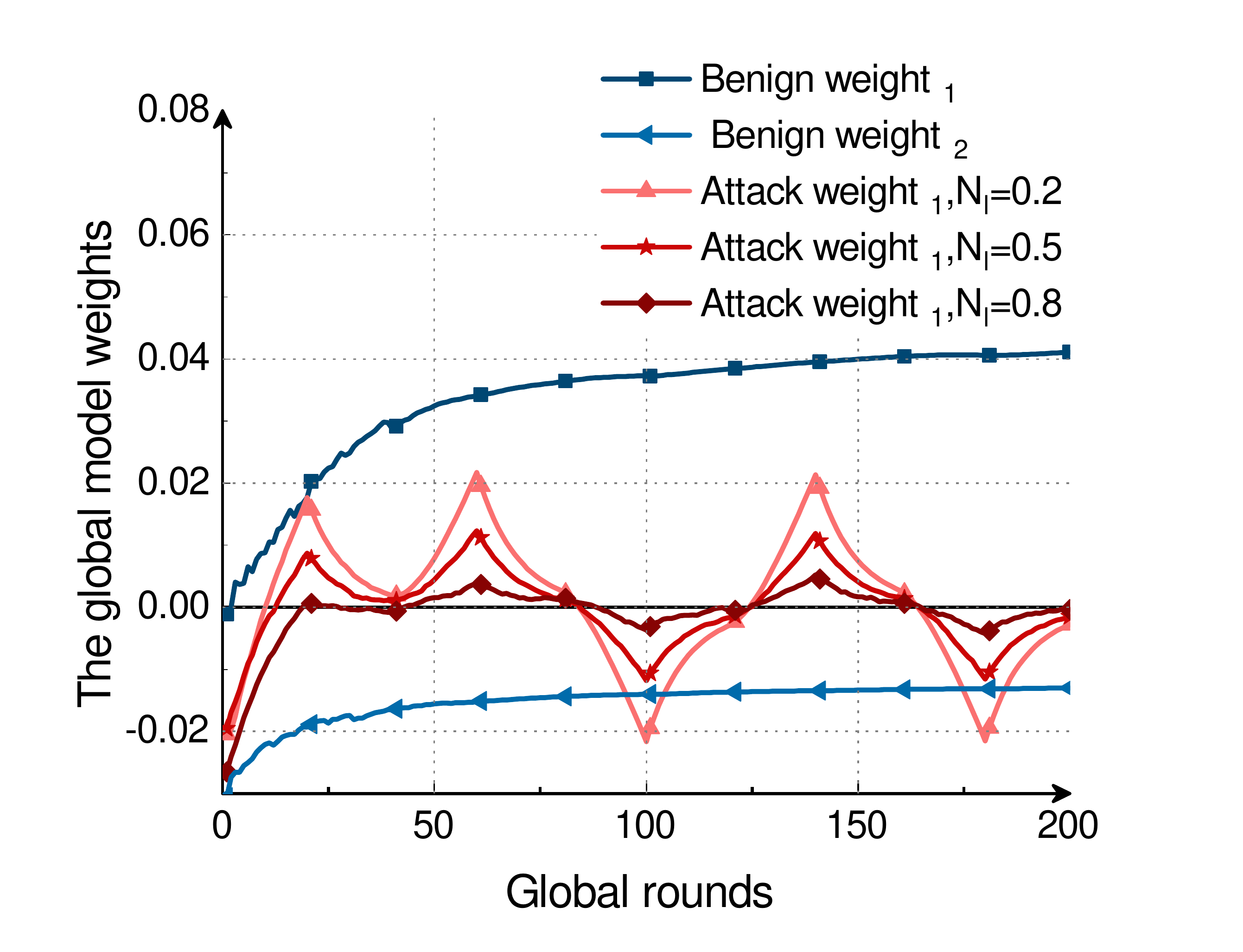}
  \caption{The influence of different noise levels on message transmission.}
  \Description{The influence of different noise levels on message transmission.}
  \label{fig8}
\end{figure}

\textbf{Our attack is robust.} Noise perturbation is a classical technique for federated learning defense. Differential privacy \cite{14,15,16} is one of the most popular methods. Differential privacy adds noise (such as Gaussian noise or Laplace noise) to the data or the model to reduce poisoning. We assume that the server can add noise to the attacker model. We experimented with the impact of different noise levels on attacker message transmission. As shown in Figure \ref{fig8}, the attacker sends messages [1,1,0,1,0] with noise levels of 10\%, 50\%, and 80\%. The high level of noise affects the attacker's transmission effect and reduces the peak signal transmission. But our method still performs well at a 50\% noise level.

\begin{figure}[h]
  \centering
  \includegraphics[width=\linewidth]{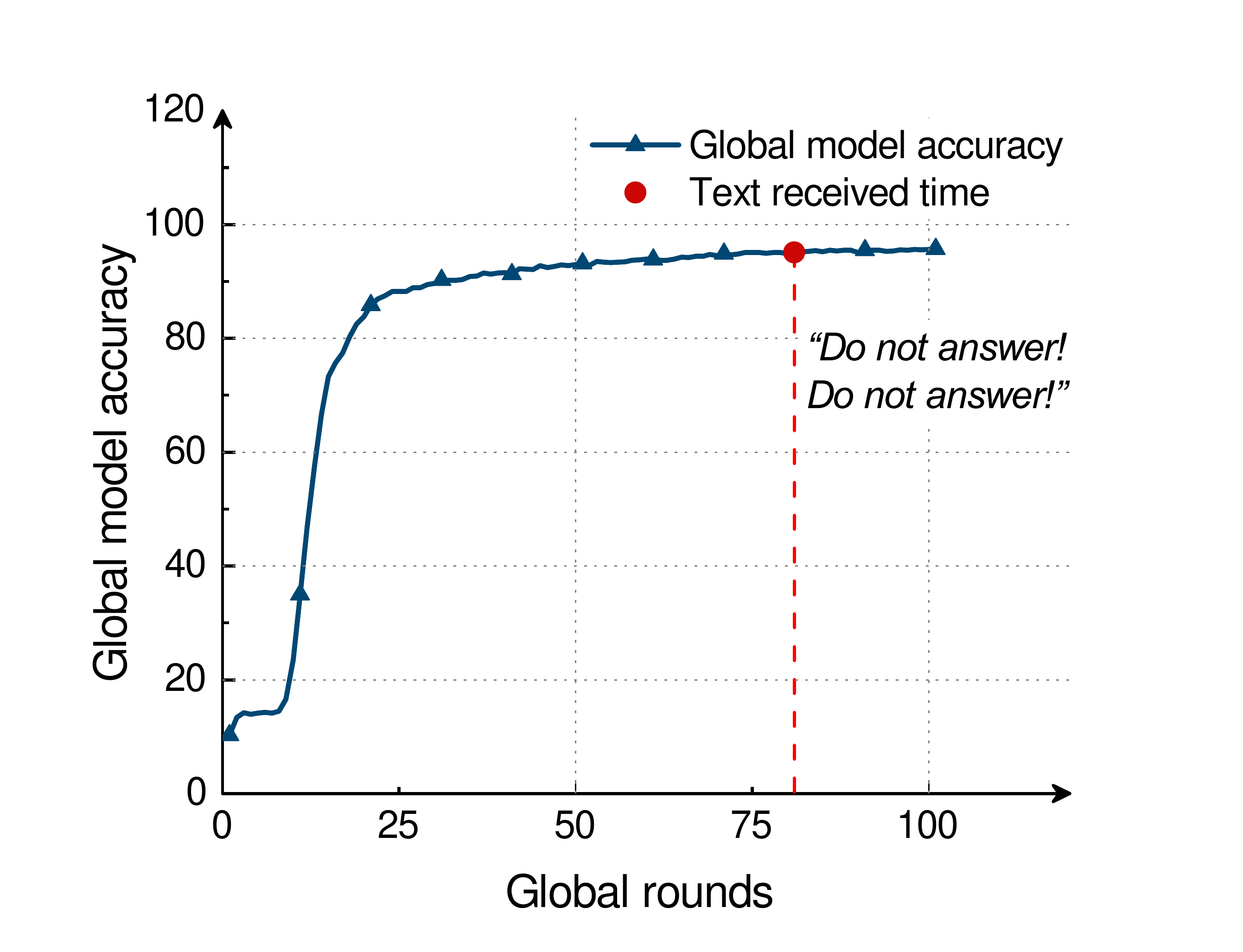}
  \caption{The time when the text message is successfully received.}
  \Description{The time when the text message is successfully received.}
  \label{fig9}
\end{figure}

\begin{figure}[h]
  \centering
  \includegraphics[width=\linewidth]{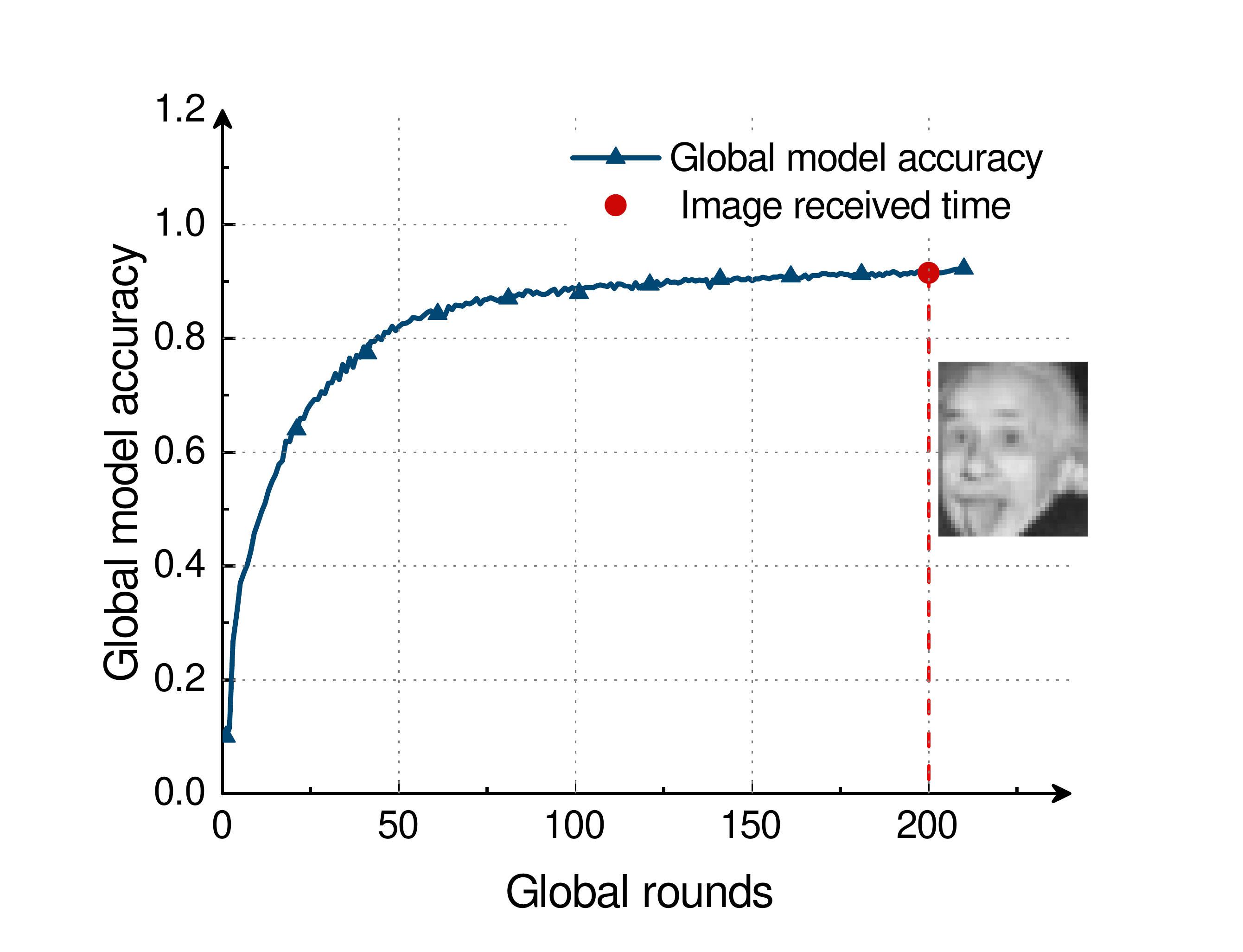}
  \caption{The time when the image message is successfully received.}
  \Description{The time when the image message is successfully received.}
  \label{fig10}
\end{figure}

\textbf{Our attack successfully delivered the messages.} We tested the message delivery capability of our method on two different networks. We show the time that the text message was successfully received in Figure \ref{fig9}. We use a 2-layer convolution neural network to send text messages. The text message is 120 bits in total and is successfully received by the receiver at the 80$th$ round. Similarly, as shown in Figure \ref{fig10}, the 11,040 bits image was successfully received at the 200th round. With the same round, we achieve more content delivery than in previous work \cite{14}. We use ResNet18 to send the image message. In the text-sending experiment, the total number of clients is 20, the sending parameter $S_i$ = 40 rounds, and the number of weights used to transmit the message is 60. In the image-sending experiment, the number of clients is 20, the sending parameter $S_i$ = 40 rounds and the number of attacker-selected weights is 2,250.

\begin{figure}[h]
  \centering
  \includegraphics[width=\linewidth]{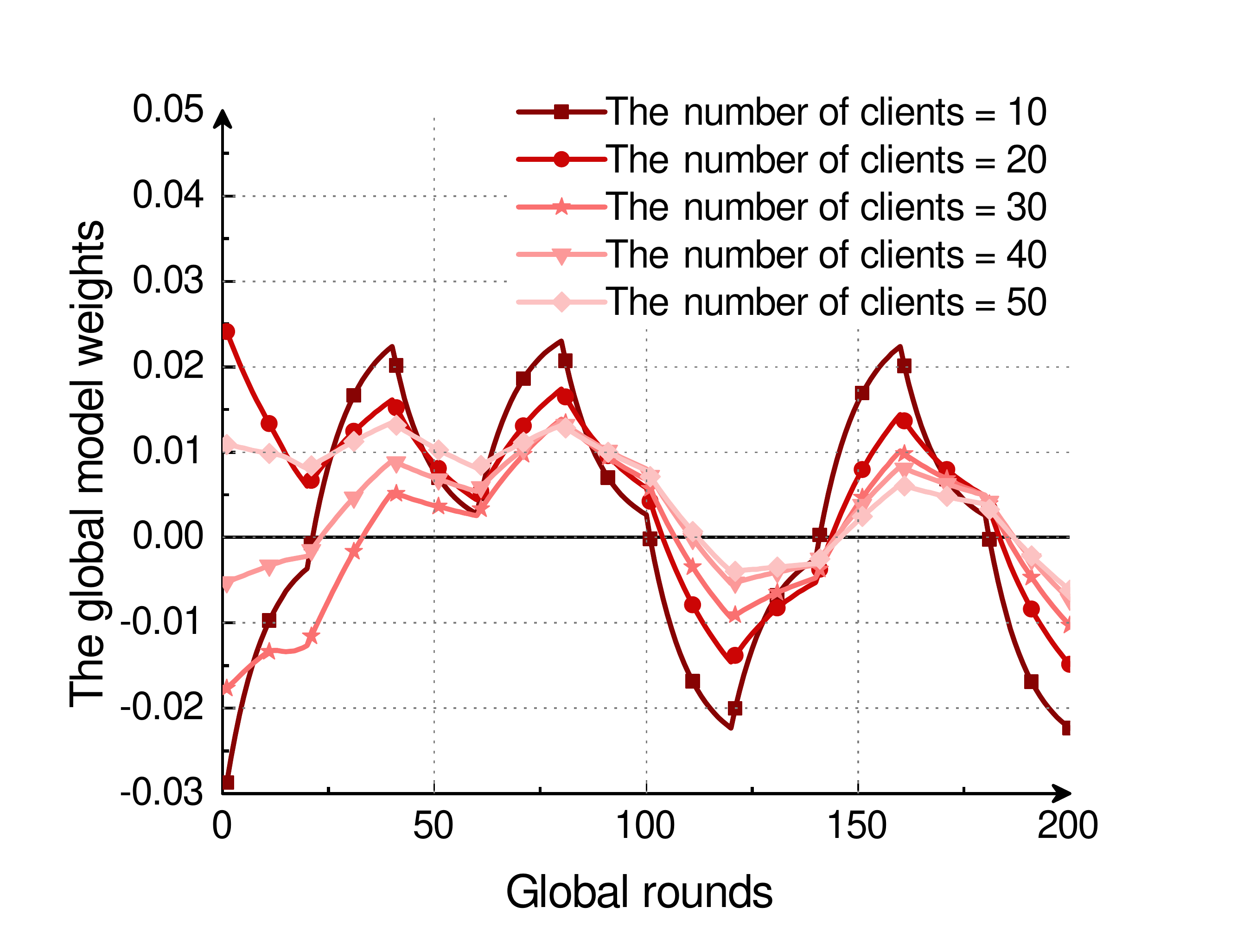}
  \caption{The time when the image message is successfully received.}
  \Description{The time when the image message is successfully received.}
  \label{fig11}
\end{figure}

\textbf{Effect of the number of clients.} As shown in Figure \ref{fig11}, we experimented with the effect of the number of clients on the transmission effect with the same above settings. There is only one message sender in the experiment. The number of clients ranges from 10 to 50, and the time to start sending messages is the 20$th$ round, sending parameters $S_i$ =40. In the first 20 rounds, the sender does back-to-zero to reduce the effect of the initial value. And, the message sent by the sender is [1, 1, 0, 1, 0]. It can be seen from Figure \ref{fig11} that the signal decreases gradually as the number of clients increases. The messages transmitted by the sender are combined with the weights of other client models through averaging. This leads to the sender's influence on the global model tough. This  can be solved by increasing the number of attackers by increasing the number of sending rounds $n$.

\section{Covert Communication Defenses}
In recent years, several defense methods have been proposed to detect and prevent covert communication attacks in federated learning such as traffic and data detection \cite{98}, and data sanitization \cite{99,100}. In this section, we will discuss the limitations of the existing defense methods for defending against covert communication attacks.

\textbf{Traffic and data detection} \cite{98}. One of the most common defense methods is traffic and data detection, which analyzes the traffic patterns and data content of the communication between clients. However, this method is not effective in defending against our attacks since we use the same number of parameters for all clients, and we disguise the smuggled data, making it indistinguishable from normal model updates.

\textbf{Data sanitization} \cite{99,100}. Data sanitization is another commonly used defense method, where noise is added to the data to destroy any toxic parts. However, our experiments show that even with high noise levels, our method is still effective in transmitting covert messages. Additionally, since the model needs to remain usable, the addition of noise is limited, making this defense method ineffective against our attacks.

\textbf{Model validation} \cite{25,39,40}. Model validation is a defense method where the server uses validation data sets to validate the accuracy of each uploaded model. However, our careful poisoning approach does not affect the accuracy of the models, making this defense method ineffective in preventing covert communication.

In light of these limitations, we propose a new defense scheme where the server records all model weight changes for all clients. This method of recording all parameters seems to work. However, this approach is computationally expensive and may require additional resources, making it less practical. Therefore, further research is necessary to develop more efficient and effective defense methods to prevent covert communication attacks in federated learning.

\section{Related Works}
\subsection{Attacks against Federated Learning}
\textbf{Disruption attacks.} Disruption attacks are the attacker aims to destroy the model's performance either partially or totally.  This attack mainly includes targeted poisoning attacks \cite{35,36,37} and untargeted poisoning attacks \cite{40}. For example, a cat and dog recognition model has a low accuracy for cats but a high rate for dogs. Or it has a low accuracy for both cat and dog recognition. This type of attack is high in destructive power and low in stealthiness.

\textbf{Controlling attacks.} The purpose of control-type attacks is to control the model output such as backdoor attacks \cite{30,31,32,33,34}. The model performs well in normal sample recognition when the input is right. However, if the input sample has a trigger (e.g., a picture of a car with a red square), the output of the model will have controlled by the attacker. For example, the model recognizes a picture of a cat as a dog when the cat picture with a backdoor trigger. This type of attack is highly stealthy and often difficult to detect.

\subsection{Defenses against Attacks}

\textbf{Perturbation-based defenses.} A perturbation mechanism is a perturbation of data or parameters to destroy the poisoned part of the data or parameters. The perturbation mechanism gains the security improvement of the model at the expense of a small amount of accuracy. One of the most famous perturbation mechanisms is differential privacy. Differential privacy mechanisms have numerous techniques in federated learning privacy preservation \cite{15,16}. Due to its rigorous mathematical theoretical foundation and good performance, there are also many related studies in security. Du et al. \cite{17} demonstrated that applying differential privacy can improve the utility of outlier detection and can effectively mitigate backdoor attacks.

\textbf{Clustering-based defenses.} Clustering-based defense is based on the premise that malicious and benign clients do not have the same purpose. This means that the statistics of the attacker and benign client model parameters will differ. Shen et al. \cite{18} propose the KMeans approach to distinguish benign clients from malicious clients. Wang et al. \cite{19} use an outlier-based detection method that can mitigate backdoor attacks. Tran et al. and Li et al. \cite{20,21} use a spectral clustering technique to identify backdoor triggers to have a higher defense against backdoor attacks.

\textbf{Comparison-based defenses.} The core of a comparison-based solution is to have a “clean” or “dirty” object for comparison. How to find the client with “clean” or “dirty” parameters is the core problem of this solution. Shen et al. \cite{22} propose a scheme to detect dirty samples by initial correctness. This approach is based on the observation that clean samples are more accurate at the initial stage of training and less accurate for samples carrying backdoor triggers. Ozdayi et al. \cite{24} find that if a client carries a backdoor task, the two updates parameter of the client will be more similar compared to other benign clients. Cao et al. \cite{25} further strengthen the assumption of federated learning by assuming that the server in federated learning has a small “clean” verification dataset. The server can use the validation dataset to test the accuracy of each client's model. The key for this type of technique to work is that the attacker's behavior is different from that of benign clients (e.g., the attacker model is less accurate).

Krum \cite{26} is a Byzantine fault-tolerant algorithm based on Euclidean distance. In the Trim-mean \cite{27} method, the server collects the parameter values from all local model updates and sorts them. This approach uses the trimmed mean as the value of the corresponding parameter in the global model update. Mhamdi et al. \cite{28} proposed Bulya, which is essentially a combination of variants of Krum and trimmed mean. Pillutla et al. \cite{29} propose methods based on the classical geometric median. These methods are effectively provided that the attacker's model statistics will be very different from those of benign clients and that the number of attackers is small.

\section{Conclusions}
This paper introduces a novel poisoning-based covert communication technique for federated learning that achieves 100\% accuracy in message transmission between two clients. We also present optimizations to increase capacity and enhance the stealthiness of the method. Extensive experiments demonstrate the effectiveness, robustness, and stealthiness of our approach. We also evaluate existing defense and detection methods, which show limited effectiveness against our covert communication attack. Our proposed scheme serves as a foundation for future federated learning attacks, such as malware hiding, and we plan to investigate methods for detecting seemingly harmless clients which can promote fairness in federated learning.

\balance

%
% The next two lines define the bibliography style to be used, and the bibliography file.
\bibliographystyle{ACM-Reference-Format}
\bibliography{references}

\end{document}